\documentclass[runningheads]{llncs}
\usepackage[T1]{fontenc}
\usepackage{graphicx}
\usepackage{cite}
\usepackage{amsmath,amssymb,amsfonts}
\usepackage[linesnumbered, ruled]{algorithm2e}
\usepackage{graphicx}
\usepackage{textcomp}
\usepackage{xcolor}

\usepackage{tabu}                     
\usepackage{multirow}                 
\usepackage{multicol}                 
\usepackage{multirow}                
\usepackage{float}                    
\usepackage{makecell}                 
\usepackage{booktabs}                 

\usepackage{stfloats}
\usepackage{hyperref}

\usepackage{etoolbox}

\usepackage{marvosym}

\begin{document}
\bibliographystyle{splncs04}

\title{CBNN: 3-Party Secure Framework for Customized Binary Neural Networks Inference}
%
%
\author{Benchang Dong \and
Zhili Chen \inst{(}\textsuperscript{\Letter}\inst{)}  \and
Xin Chen \and
Shiwen Wei \and
Jie Fu \and
Huifa Li 
}

\titlerunning{CBNN}

\authorrunning{B. Dong et al.}
%
\institute{School of Software Engineering, East China Normal University, Shanghai, China \email{zhlchen@sei.ecnu.edu.cn}}
\maketitle              
\begin{abstract}
Binarized Neural Networks (BNN) offer efficient implementations for machine learning tasks and facilitate Privacy-Preserving Machine Learning (PPML) by simplifying operations with binary values. Nevertheless, challenges persist in terms of communication and accuracy in their application scenarios.

In this work, we introduce CBNN, a three-party secure computation framework tailored for efficient BNN inference. Leveraging knowledge distillation and separable convolutions, CBNN transforms standard BNNs into MPC-friendly customized BNNs, maintaining high utility. It performs secure inference using optimized protocols for basic operations. Specifically, CBNN enhances linear operations with replicated secret sharing and MPC-friendly convolutions, while introducing a novel secure activation function to optimize non-linear operations. We demonstrate the effectiveness of CBNN by transforming and securely implementing several typical BNN models. Experimental results indicate that CBNN maintains impressive performance even after customized binarization and security measures.

\keywords{Privacy-preserving machine learning \and Secure multiparty computation \and Secure inference \and Binarized neural network \and Knowledge distillation.}
\end{abstract}
\section{Introduction}
Recently, Binarized Neural Networks (BNNs) \cite{r18} have gained popularity due to their lightweight architecture and competitive performance across various tasks. However, the model's accuracy crucially relies on the size of the training data \cite{r9}, which might need to be kept confidential under national privacy laws. To address this, Privacy-Preserving Machine Learning (PPML) technologies, leveraging cryptography techniques like Secure Multiparty Computation (MPC) and Homomorphic Encryption (HE) \cite{r2,r3,r31,r5,r6,r7,r8,r11,r13,r14}, have emerged as viable solutions.


Binarization is a well-established approach for optimizing inference processes. This study focuses specifically on secure inference using BNNs. In contrast to regular Neural Networks (NNs), BNNs offer a simpler computational framework due to their binary-valued model parameters and activations (-1 and 1). Prior works \cite{r15,r12} have identified BNNs' suitability for high-precision security applications, demonstrating BNN inference with techniques such as MPC and HE. For instance, the XONN framework \cite{r12} optimizes BNN multiplication through XOR and bit-count operations. However, the underlying MPC techniques in XONN, a 2-party framework, restrict its performance. Some subsequent 3-party frameworks \cite{r15} based on secret sharing face challenges in efficiently implementing bit-count, often resorting to the same network, potentially leading to increased communication costs and accuracy losses. To enhance the performance of secure BNN inference, we refine the architecture and training methods of traditional BNN model, referred to as an "MPC-Friendly" model. This model exhibits superior performance characteristics. Additionally, we develop tailored secure protocols for the MPC-friendly model.

In this study, we introduce CBNN, a three-party secure computation framework tailored for customizing and evaluating BNNs. This framework, secure against semi-honest adversaries, divides participants into three roles: data owner, model owner, and helper. For customization, CBNN transforms the standard BNN, owned by the model owner, into an MPC-friendly BNN using specialized architecture and training methods. For evaluation, CBNN devises secure computation protocols for various model layers. Typically, the inference input remains confidential with the data owner, while the neural network's weights and biases are kept secret by the model owner. Upon completion of the evaluation, the framework safely discloses the genuine inference result to the data owner.

Our contributions can be summarized as follows:

\begin{itemize}
\item {}\textbf{Introducing CBNN}, an innovative 3-party secure computation framework tailored for customized BNN inference. Through cutting-edge methods, we are the first to transform any standard BNN model into an MPC-compatible BNN model, enabling secure inference without compromising performance.

\item {}\textbf{Introducing novel customized training methods} specifically designed for MPC-friendly BNNs. When trained on standard datasets, CBNN-tailored BNNs exhibit superior performance over prior approaches. We integrate knowledge distillation and separable convolution to boost inference accuracy and minimize model parameters, maintaining the original architecture.

\item {}\textbf{Designing Secure Protocols} for Basic Inference. CBNN introduces an innovative MSB extraction method to optimize non-linear computations. Leveraging 3-party oblivious transfer, we implement activation functions for enhanced secret sharing in secure DNN inference. Additionally, we refine maxpooling layers and introduce an adaptive fusing BN protocol to reduce costs.

\item {}\textbf{Demonstrating Applicability and Practicality} through Implementation on Diverse Datasets. To assess the impact of customization methods on inference accuracy and computation/communication costs of customized BNNs, CBNN transforms typical BNN models trained on standard datasets to enable secure inference. Experimental results demonstrate the applicability and practicality of CBNN.

\end{itemize}

\section{Preliminaries}

\subsection{Binarized Neural Networks}
BNN \cite{r18} is a subtype of Neural Networks, where weights and activations are limited to the values \{-1,1\}. 
Below, we outline the neural network layer's functionalities.

{\bfseries Linear Layers:\ } In BNNs, linear operations are executed through Fully-Connected (FC) and Convolution (CONV) layers. 

{\bfseries Activation:} In BNNs, the non-linear operation primarily serves to binarize the normalized value. This study specifically employs the Sign and ReLU functions for hidden layers, which are implemented through the extraction of the most significant bit (MSB).

{\bfseries Batch Normalization:} Batch Normalization (BN) layers are typically employed after linear layers to normalize their outputs. This normalization involves subtracting $\beta$ and dividing by $\gamma$, which are trainable parameters.

{\bfseries MaxPooling:} The maxpooling layer slides a window over the image channels, aggregating elements within the window into a single output element. This typically reduces data size without altering the number of channels. 

\subsection{Knowledge Distillation}
\begin{figure}[htbp]
\centering
\includegraphics[width=0.75\linewidth]{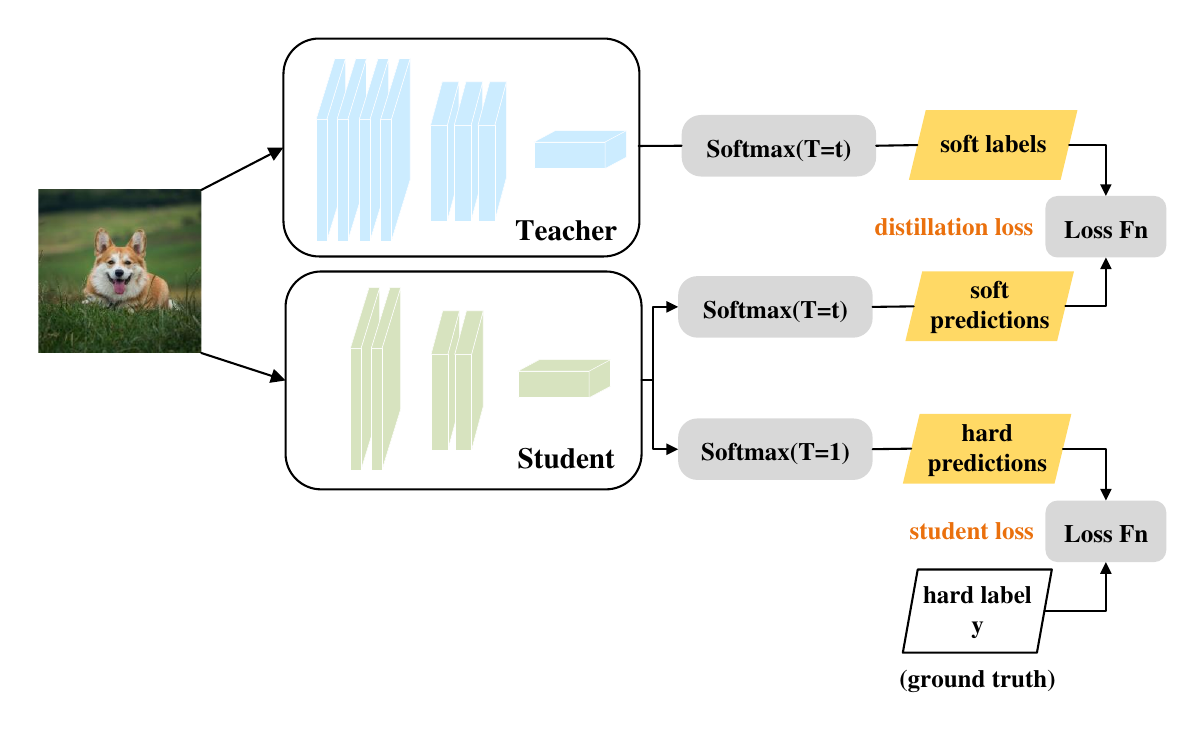}
\caption{Knowledge distillation setup involves a teacher network that boasts high precision and a student network that operates as a low-precision binary network.}
\label{fig:1}
\end{figure}

The standard approach in Knowledge Distillation (KD) \cite{r19} employs a teacher-student paradigm, where a large and complex teacher network imparts knowledge to a smaller student network trained on the identical dataset. Figure \ref{fig:1} illustrates the entire KD algorithm's workflow.

{\bfseries Distillation:} DNNs typically convert the layer output, known as logit $z_i$, into a class probability $q^T_i$ for each class using the softmax function.

\begin{equation}
  q^T_i=\frac{exp(z_i/T)}{\sum_jexp(z_j/T)}
  \label{formula:1}
\end{equation}

{\color{black} The concept of temperature $T$, introduced by Hinton \cite{r19}, determines the softness of the probability distribution across classes.

}

{\bfseries Teacher-Student Network:\ }
During training, as depicted in Figure \ref{fig:1}, the computation of the student model's loss function follows the standard protocol. The dataset comprises inputs and labels, with the label of the ground truth referred to as the hard label. The student loss is determined using the cross-entropy loss function $H$.

\begin{equation}
  H(p,q)=-\sum_{i=1}^np_ilog(q_i)
\end{equation}

The term $class$ denotes the accurate outcome of the classification process. Depending on the hard label $y_i$, if $y_i=1$ when $i$ corresponds to the correct class, or $y_i=0$ when $i$ does not match the correct class, the student loss can be simplified as follows:

\begin{equation}
  H_{stu}(y,q)=-log(q_{class})
  \label{formula:3}
\end{equation}
As Figure \ref{fig:1} illustrates, the teacher model employs the softmax function under a specific temperature $T$ to generate predictions $q^T_i$, referred to as soft labels. Given the identical temperature $T$ and inputs, the student model computes predictions $p^T_i$, known as soft predictions. Following the formula \ref{formula:3}, the teacher loss is defined as:

\begin{equation}
  H_{tea}(p^T,q^T)=-\sum_{i=1}^np^T_ilog(q^T_i)
  \label{formula:4}
\end{equation}

During the training process, the overall loss function $L$ is computed as a combination of the student loss and the teacher loss.

\begin{equation}
 L(x,y) = \lambda H_{stu}(y,q) + (1 - \lambda) H_{tea}(p^T,q^T)
 \label{formula:5}
\end{equation}

The essence of knowledge distillation lies in the utilization of a higher temperature $T$ to generate soft labels in the teacher model, facilitating the transfer of knowledge. The weighting factor $\lambda$ serves to prioritize the output of a specific loss function relative to others. Additionally, $y$ denotes the hard label corresponding to the soft label $p^T$. By training the student model with both soft and hard labels, it can achieve comparable accuracy to the more complex teacher model.

\subsection{Replicated Secret Sharing}

Secrets can be securely distributed among multiple parties through Secret Sharing (SS) schemes\cite{r21,r22}, ensuring zero privacy breaches. 
In our study, we adopt the Replicated Secret Sharing (RSS) scheme proposed by Araki et al\cite{r20} to minimize communication overhead and boost computational efficiency.

The secret value $x$, residing in the ring $\mathbb{Z}_{2^{k}}$ and represented as $x=x_{0}+x_{1}+x_{2}$, is distributed among three participants. In this setup, $P_i$ possesses the shares $(x_0,x_1)$. We denote the RSS share within an arithmetic circuit as $\left[x\right]_3^A$. The reconstruction of these shares can be achieved by any two participants through the operation: $ x=x_{0}+x_{1}+x_{2}\mod2^{k}$.

For the addition of secret shares, denoted as $\left[x+y\right]_3^A$ (or the addition of secret shares with a constant value, represented as $\left[x+c\right]_3^A$), participants simply need to perform a local addition of their respective shares. Specifically, this involves adding $\left[x\right]_3^A$ and $\left[y\right]_3^A$ to obtain $(x_i+y_i,x_{i+1}+y_{i+1})$ or adding $\left[x\right]_3^A$ with a constant $c$ to yield $(x_i+c,x_{i+1})$.

For the multiplication of secret shares, denoted as $\left[z\right]_3^A=\left[xy\right]_3^A$, the RSS scheme utilizes a randomness $a_0+a_1+a_2=0 \mod 2^k$ to generate an additive mask. Each participant $P_i$ possesses $z_i$ which can be interpreted as: $z_i=x_iy_i+x_iy_{i+1}+x_{i+1}y_i+a_i$ and $z=\sum_{i=0}^2z_i \mod 2^k$. To adhere to the RSS scheme, the protocol requires a re-sharing process where $P_i$ sends $z_i$ to $P_{i-1}$. Finally, each participant $P_i$ holds the RSS share $\left[z\right]_3^A=(z_i,z_{i+1})$.

\subsection{Oblivious Transfer}

\begin{algorithm}\small
\label{alg:1}
\SetKwInOut{Input}{input}\SetKwInOut{Output}{output}

\Input{Sender inputs $(m_0,m_1)$, Receiver and Helper input choose bit $c\in\mathbb{Z}_{2}$.}
\Output{Receiver outputs $m_c$, Sender and Helper output $\perp$.}
\BlankLine
\BlankLine
  Sender and Receiver generate common random bit string $mask_0,mask_1$\;
  Sender masks the message $s_i=m_i\oplus mask_i,i\in \{0,1\}$\;
  Sender sends $(s_0,s_1)$ to Helper\;
  Helper selects and sends the message $s_c$ to Receiver\;
  Receiver unmasks the message $m_c=s_c\oplus mask_c$.
  
  \caption{Three-Party Oblivious Transfer}
\end{algorithm}

The Oblivious Transfer (OT) protocol, a cornerstone of secure computation, involves two parties: a sender and a receiver\cite{r23,r24}. 

Building upon the traditional 1-out-of-2 OT protocol, this work introduces a third party, referred to as a helper \cite{r15}.
The ideal function for the three-party OT can be expressed as: $((m_0,m_1),c,c)\mapsto (\perp,m_c,\perp)$. This work utilizes replicated secret sharing and implements an efficient three-party OT protocol \ref{alg:1} for the subsequent secure activation function. The security of the scheme is ensured as the shares are typically held by the participants, satisfying the $(2,3)$ threshold.

\section{The CBNN Framework}
\begin{figure}[htbp]
\centering
\includegraphics[width=0.7\linewidth]{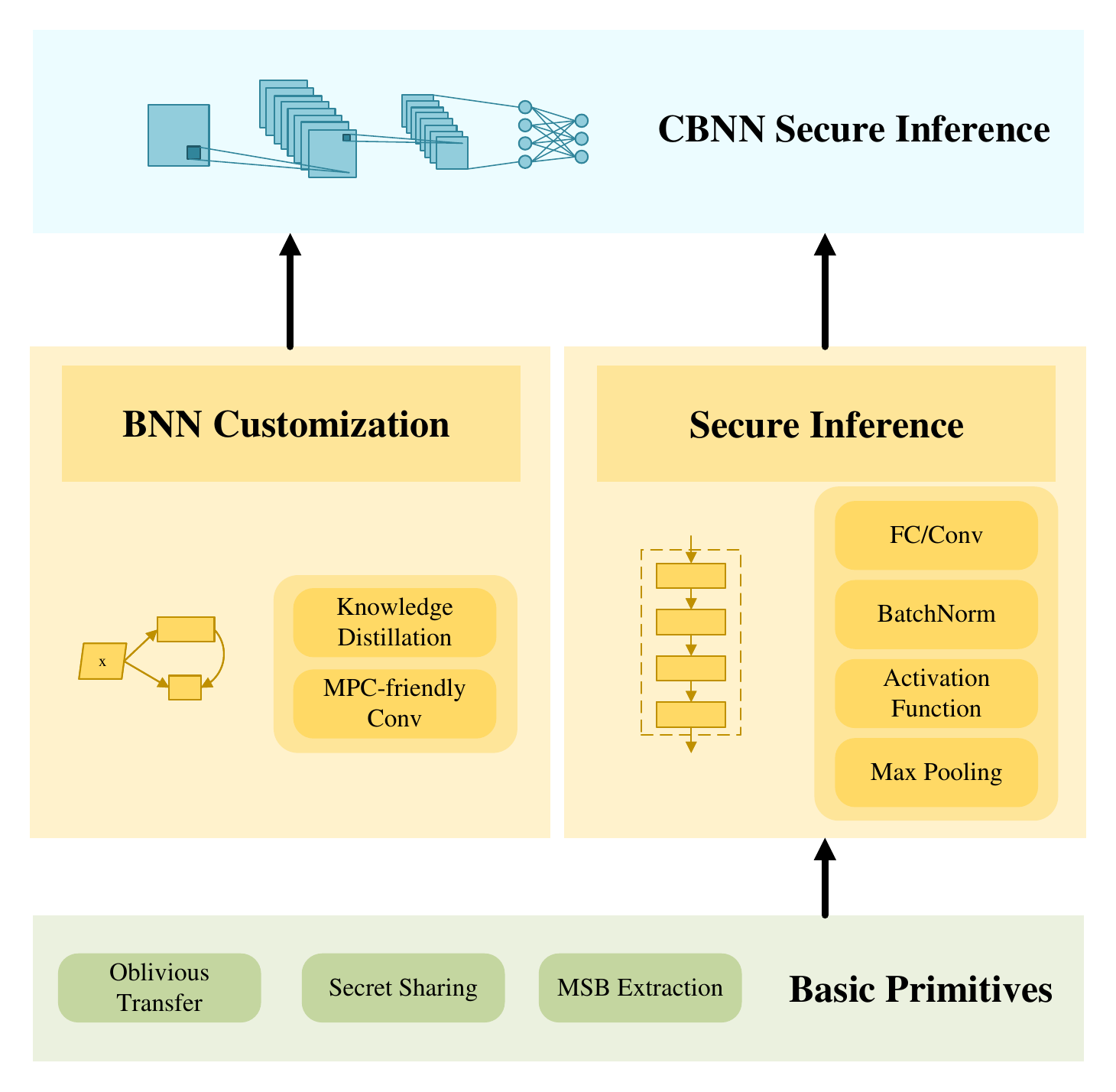}
\caption{The architecture of CBNN, including the BNN customization method and secure inference protocols with basic primitives.}
\label{fig:2}
\end{figure}

{\color{black}
Figure \ref{fig:2} outlines the architecture of CBNN, encompassing two primary components: BNN customization and Secure Inference. The customization phase strives to develop an accurate and MPC-compatible BNN by leveraging knowledge distillation and MPC-friendly convolutions during training. Secure inference, on the other hand, builds upon the customized BNNs, optimizing secure protocols for crucial operations such as activation functions.

}

\subsection{Customized Binary Neural Networks}
This framework aims to customize BNNs for secure inference using established networks tailored for renowned datasets. It transforms standard networks into MPC-compatible models, considering the implications of MPC. 

\textbf{Customized binarization:}
In typical BNNs, activations and weights are binarized to {-1, +1}. Prior works replaced expensive matrix multiplication with XNOR operations in MPC schemes like Garbled Circuits. This study opts for replicated secret sharing as the MPC framework. {\color{black}We partially implement binarization in the activation layer, favoring Sign functions, with ReLU occasionally employed to boost scalability and adaptability}. Nevertheless, full-precision (32 bits) model parameters are maintained to ensure high inference accuracy.

\textbf{MPC-friendly convolutions:}
{\color{black} By substituting CNN convolutions with MPC-compatible alternatives, we have effectively mitigated computational overhead}. 

\begin{figure}[htbp]
\centering
\includegraphics[width=0.7\linewidth]{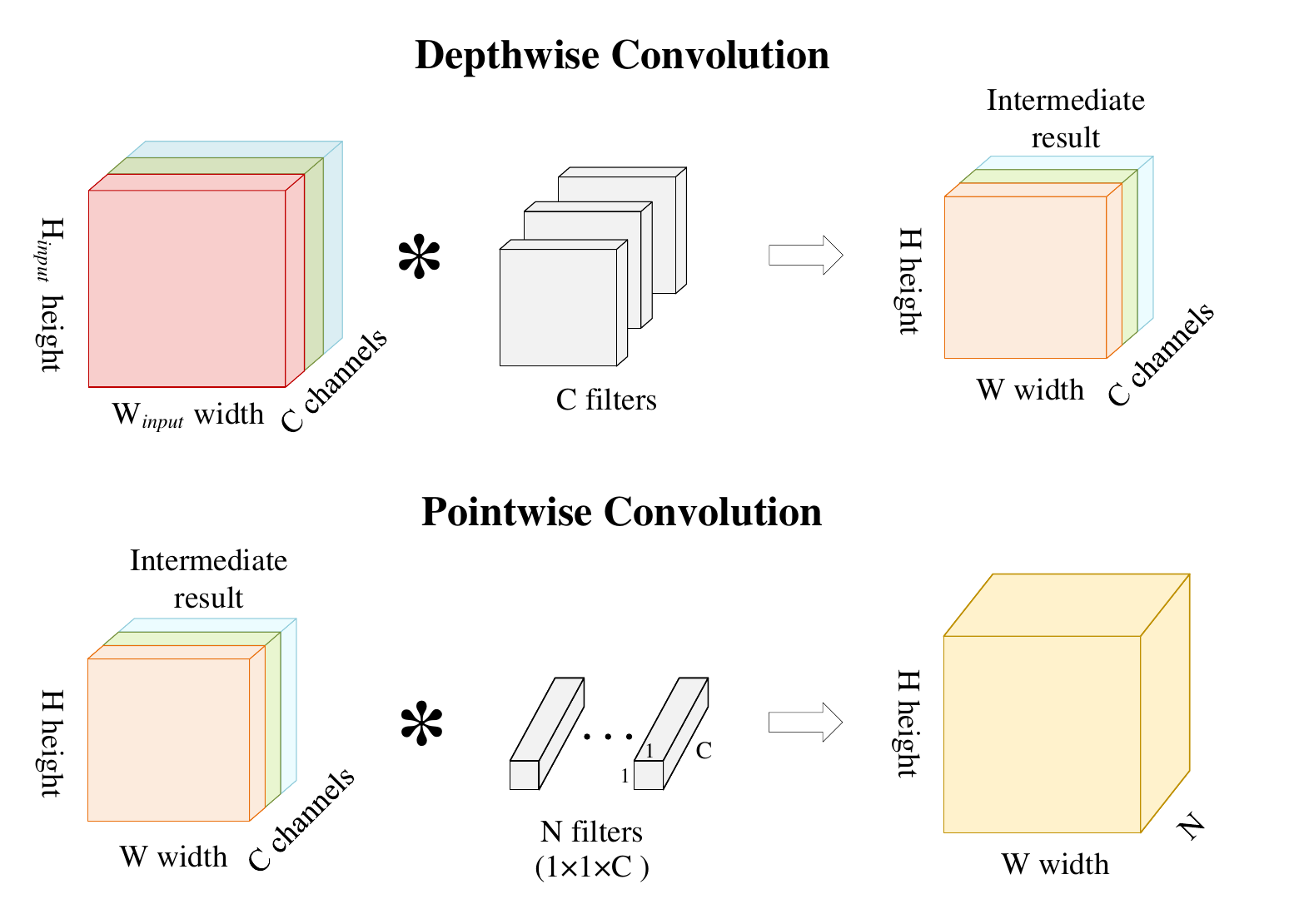}
\caption{ Workflow of MPC-friendly convolution. MPC-friendly convolution is implemented by Separable convolution, which consists of two steps, namely, Depthwise Convolution and Pointwise Convolution.}
\label{fig:3}
\end{figure}

Figure \ref{fig:3} depicts the traditional convolution as a two-step process: Depthwise and Pointwise convolutions, with Algorithm \ref{alg:2} executed twice. High network delay increases inference time due to more communication rounds, which cannot be offset by reduced communication cost alone. Falcon\cite{r10} has shown that as networks grow, computation dominates runtime. In summary, MPC-friendly convolutions are ideal for complex networks with low network latency.

\textbf{Customized training via Knowledge distillation:}
During training, the KD algorithm is employed to transfer knowledge from a complex network to a customized low-precision network with binarization.
In this study, we introduce a Customized BNN as the student network, incorporating binarized activation and MPC-friendly convolutions.
Experimental results demonstrate that the distilled network exhibits superior inference performance, both in ciphertext and plaintext.

In summary, Customized binarization and MPC-friendly convolutions have shaped the structure and parameters of an MPC-compatible model. This framework incorporates the KD algorithm, enabling the training of a high-precision, MPC-friendly model leveraging a general high-precision model.
Next, we will detail the implementation of various layers in customized BNN secure inference.

{\color{black}
\subsection{Basic Primitives  For Secure Inference}
Before delving into the specifics of each layer protocol, we outline the underlying principles that facilitate the implementation of secure inference protocols based on fundamental primitives.
}

In the CBNN setting, there are three participants: the data owner ($P_0$), the model owner ($P_1$), and the helper ($P_2$). For a given participant $P_i$, the subsequent and preceding participants are denoted as $P_{i+1}$ and $P_{i-1}$, respectively, with indices wrapped around using the modulo $3$ operation.

\textbf{Correlated Randomness:} This work utilizes PRFs (Pseudo-Random Functions) to produce random values, denoted as $F(k,cnt)$. Each participant $P_i$ and its subsequent participant $P_{i+1}$ share a secret random seed $k_i$. Specifically, $P_i$ possesses $(k_i,k_{i+1})$ for $i = 0, 1, 2$. The counter $cnt$ is incremented upon each invocation to ensure uniqueness.

{\color{black}
\textbf{3-out-of-3 randomness:\ } 
$P_i$ computes $a_i=F(k_{i+1},cnt)-F(k_{i},cnt)\mod 2^l$ as its share, and all participants generate an additive secret sharing of $0$ due to that $\sum_{i=0}^2a_i\equiv 0\mod 2^l$. 

\textbf{2-out-of-3 randomness:} 
$P_i$ computes $(a_i,a_{i+1})=(F(k_i,cnt),F(k_{i+1},cnt))$ as its share, and all participants generate a RSS of a random number $a$ due to $\sum_{i=0}^2a_i\equiv a\mod 2^l$.  
}

\textbf{Encoding and Secret sharing:\ }
Most prior works in secure inference rely on fixed-point arithmetic, necessitating the encoding of all model parameters and inputs into $\mathbb{Z}_{2^l}$ with a secret sharing scheme. Typically, $l$ is set to $32$.

For our customized BNN, we designate the input, output, and weights of each layer as $X$, $Z$, and $W$ respectively. These values are encoded into an integer ring and then secretly shared among participants.

In the CBNN context, for a random plaintext $x$, arithmetic shares $\left[x\right]_3^A$ represent 2-out-of-3 replicated secret sharing modulo $2^l$ where $P_i$ possesses $(x_i,x_{i+1})$. Reconstruction of shares requires two participants: $ x=x_{0}+x_{1}+x_{2}\mod2^{l}$. For a random bit $y$ in plaintext, binary shares $\left[y\right]_3^B$ denote 2-out-of-3 replicated secret sharing modulo $2$. 


\begin{algorithm}\small
\label{alg:2}
\SetKwInOut{Input}{input}\SetKwInOut{Output}{output}

\Input{$P_i$ inputs $(W_i,W_{i+1}),(X_i,X_{i+1}),(b_i,b_{i+1})$.}
\Output{$P_i$ outputs $(Z_i,Z_{i+1})$, shares of Linear layer output.}
\BlankLine
\BlankLine
  $P_i$ invokes 3-out-of-3 randomness $a_i,\sum_{i=0}^2a_i\equiv 0\mod 2^l$\;
  $P_i$ computes locally $Z_i=matmul(W_i,X_i)+matmul(W_{i+1},X_i)+matmul(W_i,X_{i+1})$ or $Conv(W_i,X_i)+Conv(W_{i+1},X_i)+Conv(W_i,X_{i+1})$\;
  $P_i$ computes $Z_i=Z_i+b_i+a_i$\;
  Reshare: $P_i$ sends $Z_i$ to $P_{i-1}$\;
  $P_i$ holds $(Z_i,Z_{i+1})$

  \caption{Linear Layer Inference}
\end{algorithm}

\subsection{Linear Operations}

This study introduces an enhanced and less interactive linear layer inference protocol leveraging the RSS scheme. 
This approach extends multiplication to Matrix Multiplication (for FC layers) and Convolution (for CONV layers) as detailed in Algorithm \ref{alg:2}.

\textbf{Truncation:\ }
Due to fixed-point arithmetic encoding in secret computation, the outputs $Z_i$ of the linear layer must be truncated to maintain accuracy after multiplying two fixed-point numbers. This work utilizes the truncation protocol $\Pi_{\mathsf{trunc}1}$ from \cite{r5}, which requires two rounds of communication.

\textbf{Share Conversion:} In the subsequent Non-Linear portions of the model (e.g., Sign activation), converting between various share formats is crucial. 

The conversion from $\left[x\right]_3^B$ to $\left[x\right]_3^A$. In the Three-Party Oblivious Transfer protocol \ref{alg:1}, the data owner fulfills the role of the sender, the model owner serves as the receiver, and the helper acts as the helper. Within the RSS scheme, each participant possesses shares $(x_i^B,x_{i+1}^B)$. The model owner initiates by generating random values jointly with the other participants $(x_1,x_2)$ using 2-out-of-3 randomness. The data owner and helper contribute the choice bit $x_0^B\in Z_2$, while the model owner constructs the message $(m_0,m_1)$ with $m_i = (i\oplus x_1^B\oplus x_2^B) - x_1 - x_2$ to ensure the data owner can reconstruct the arithmetic secret shares $x_0$ using the choice bit $x_0^B$.

For the conversion from $\left[x\right]_3^A$ to $\left[x\right]_3^B$, CBNN implements this transformation within the MSB Extraction protocol \ref{alg:3}.

\begin{algorithm}\small
\label{alg:3}
\SetKwInOut{Input}{input}\SetKwInOut{Output}{output}

\Input{$P_0,P_1,P_2$ hold secret sharing of $x\in\mathbb{Z}_{2^l}$}
\Output{All parties get shares of most significant bit $\ MSB\in\mathbb{Z}_2$}
\BlankLine
\BlankLine
  $P_i$ invokes 2-out-of-3 randomness, generates shares of a random, private bit$\left[\beta\right]_3^B$, $\beta\in\mathbb{Z}_2$, and shares of integer $r\in\mathbb{Z}_2^{l-1}$\;
  Convert shares $\left[\beta\right]_3^B$ to $\left[\beta\right]_3^A$\;
  $P_1$ generates random, private values $\alpha_{1},\alpha_{2}$ and sends $\alpha_{2}$ to $P_{2}$\;
  Run 3-party-OT protocol\;
  $P_0$ inputs choose bit $\beta_0^B$\;
  $P_2$ inputs choose bit $\beta_0^B$\;
  $P_1$ construct $(m_0,m_1)$, $m_i = (i\oplus \beta_1^B\oplus \beta_2^B) - \alpha_1 - \alpha_2(\mod 2^l)$\;
  Parties get arithmetic shares $\left[\beta\right]_3^A$ masked by $\alpha_{1},\alpha_{2}$\;
  Compute shares $\left[u\right]_3^A=\left[(-1)^\beta xr\right]_3^A$\;
  reveal shares of $u$ and compare with $2^{l-1}$
  let $\beta'=1$ if$(u>2^{l-1})$ and $\beta'=0$ otherwise\;
  return Shares of $\beta'\oplus\beta\in\mathbb{Z}_2$

  \caption{MSB Extraction}
\end{algorithm}

\subsection{Activation Function}
{\color{black}
Having finished performing linear layer inference, participants proceed with the activation layer.
}

\textbf{MSB Extraction:} To securely implement activation functions like Sign or ReLU, the proposed Most Significant Bit (MSB) Extraction protocol builds on the optimized approach from Algorithm 1 in \cite{r10}. {\color{black}Distinct from prior works, this algorithm forgoes costly bit decomposition and achieves extraction through the utilization of an arithmetic share-based mask.} This protocol effectively extracts the MSB of $x$ where the participants jointly hold shares in $\mathbb{Z}_{2^l}$.

This protocol \ref{alg:3} efficiently implements MSB Extraction with minimal communication overhead and optimal computational performance. Both SecureNN \cite{r7} and Falcon \cite{r10} adopt secure comparison operations to assess the original inputs against zero for MSB extraction. We streamline these private comparison protocols, eliminating the dependency on bit decomposition.

\begin{algorithm}\small
\label{alg:4}
\SetKwInOut{Input}{input}\SetKwInOut{Output}{output}

\Input{$P_0,P_1,P_2$ inputs Binary secret shares $\left[MSB(x)\right]_3^B$}
\Output{All parties get shares $\left[Sign(x)\right]_3^A=\left[1\oplus MSB(x)\right]_3^A$}
\BlankLine
\BlankLine
  $P_1$ generates and distributes random, private values $\beta_{1},\beta_{2}$, $P_0$ holds $\beta_{1}$, $P_1$ holds $(\beta_{1},\beta_{2})$, $P_2$ holds $\beta_{2}$\;
  Run 3-party-OT protocol\;
  $P_0$ inputs choose bit $MSB_0^B$\;
  $P_2$ inputs choose bit $MSB_0^B$\;
  $P_1$ construct $(m_0,m_1)$, $m_i=(1\oplus i\oplus MSB_1^B\oplus MSB_2^B)-\beta_1-\beta_2\mod 2^l$\;
  $P_0$ gets and sends $m_{c} = (1\oplus MSB)-\beta_1-\beta_2$ to $P_2$\;
  Parties get shares of $(1\oplus MSB)$ masked by $\beta_1,\beta_2$\;
  return Shares of $Sign(x)\in\mathbb{Z}_{2^l}$

  \caption{Secure Sign}
\end{algorithm}

\textbf{Sign Function:} To mitigate accuracy losses associated with traditional binary networks, we utilize both the $Sign$ and $ReLU$ functions as activation functions in our customized binary neural network. In previous work (XONN), the $Sign$ function enhances neural network nonlinearity, serving as the activation function and binarizing the linear layer's output as the foundation of BNNs. Notably, the secure $Sign$ function \ref{alg:4} can be efficiently implemented using the MSB Extraction protocol \ref{alg:3}, with parties possessing the binary secret shares $\left[MSB(x)\right]_3^B$. The secure $Sign$ protocol is essentially conversion between binary and arithmetic shares.

\begin{algorithm}\small
\label{alg:5}
\SetKwInOut{Input}{input}\SetKwInOut{Output}{output}

\Input{$P_0,P_1,P_2$ inputs Arithmetic secret shares $\left[x\right]_3^A$ and Binary secret shares $\left[MSB(x)\right]_3^B$}
\Output{All parties get shares $\left[ReLU(x)\right]_3^A=\left[(1\oplus MSB(x))*x\right]_3^A$}
\BlankLine
\BlankLine
  $P_1$ and $P_0$ generates and distributes random, private values $(\alpha_{1},\alpha_{2}),(\gamma_0,\gamma_1)$ respectively\;
  Run 3-party-OT protocol\;
  \textbf{$P_0$} inputs choose bit $MSB_0^B$\;
  \textbf{$P_2$} inputs choose bit $MSB_0^B$\;
  \textbf{$P_1$} construct $(m_0,m_1)$, $m_i=(1\oplus i\oplus MSB_1^B\oplus MSB_2^B)*(x_1+x_2)-\alpha_{1}-\alpha_{2}$\;
  Parties get shares of $(1\oplus MSB)*(x_1+x_2)$ masked by $\alpha_{1},\alpha_{2}$\;
  \textbf{Data owner $P_0$} and \textbf{Model owner $P_2$} switch roles in OT protocol, run 3-party-OT protocol again\;
  \textbf{$P_1$} inputs choose bit $MSB_2^B$\;
  \textbf{$P_2$} inputs choose bit $MSB_2^B$\;
  \textbf{$P_0$} construct $(m_0,m_1)$, $m_i=(1\oplus i\oplus MSB_0^B\oplus MSB_1^B)*x_0-\gamma_0-\gamma_1$\;
  Parties get shares of $(1\oplus MSB)*x_0$ masked by $\gamma_0,\gamma_1$\;
  \textbf{$P_i$} computes shares locally: $\left[(1\oplus MSB(x))*x\right]_3^A = \left[(1\oplus MSB(x))*(x_0+x_1+x_2)\mod 2^l\right]_3^A$\;
  return Shares of $ReLU(x)\in\mathbb{Z}_{2^l}$

  \caption{Secure ReLU}
\end{algorithm}

\textbf{ReLU Function:} In conventional secure computation protocols, implementing secure $ReLU$ typically requires a comparison with the value $0$, which necessitates a conversion operation. However, this work introduces an efficient $ReLU$ protocol \ref{alg:5} leveraging the MSB Extraction Protocol~\ref{alg:3} and the 3-party-OT Protocol~\ref{alg:1}. Parties provide the original linear layer outputs, arithmetic secret shares $\left[x\right]_3^A$, and binary secret shares $\left[MSB(x)\right]_3^B$. The objective is to obtain shares of the activation result $\left[ReLU(x)\right]_3^A$ among the parties.

\subsection{Batch Normalization}
Numerous models employ the Batch Normalization layer to enhance generalization capabilities. {\color{black}Prior research has focused on enhancing model inference efficiency through BN layer optimization. This study introduces an adaptive fusing layers protocol tailored to distinct activation functions.}

The BN layer can be formulated as:

\begin{equation}
\label{formula:6}
  Y=\gamma\frac{X-\mu}{\sqrt{\sigma^2+\epsilon}}+\beta
\end{equation}

The values $\mu$ and $\sigma^2$ represent the mean and variance computed across a batch, respectively. The term $\epsilon$ is a small constant added for numerical stability, while $\gamma$ and $\beta$ denote the scaling and shift factors, respectively. Alternatively, Formula \ref{formula:6} can be expressed as:

\begin{equation}
  \label{formula:7}
  Y=\frac\gamma{\sqrt{\sigma^2+\epsilon}}X+\beta-\frac{\gamma\mu}{\sqrt{\sigma^2+\epsilon}}
\end{equation}

BN layers usually follow the Linear layers, as the activation layers follow BN layers. 

\textbf{BN with Sign function:\ }In the scenario of activation function $Sign$, The activation layer after BN layer can be interpreted as:

\begin{equation}
  Sign(\frac\gamma{\sqrt{\sigma^2+\epsilon}}x+\beta-\frac{\gamma\mu}{\sqrt{\sigma^2+\epsilon}})=Sign(\gamma^{\prime}x+\beta^{\prime})
  \label{formula:8}
\end{equation}

In formula \ref{formula:8}, $\gamma^{\prime}=\frac\gamma{\sqrt{\sigma^2+\epsilon}},\beta^{\prime}=\beta-\frac{\gamma\mu}{\sqrt{\sigma^2+\epsilon}}$, and $\gamma^{\prime}$ is a positive value. The fusing protocol between the activation and BN layer are implemented with $Sign(x+\frac{\beta'}{\gamma'})$. In this case, the model owner needs to split $\frac{\beta'}{\gamma'}$ into arithmetic shares, and then add $\left[\frac{\beta'}{\gamma'}\right]_3^A$ to the output share of the linear layer in the pre-processing phase. 

\textbf{BN with ReLU function:\ } In the context of ReLU activation, the linear layer's operations are simplified as:

\begin{equation}
  Z=matmul/Conv(W,X)+b
\end{equation}

The BN layer and linear layer are integrated into a unified linear layer, with parameters $W$ (weight) and $b$ (bias) defined as follows:

\begin{equation}
  W = W_{FC}\cdot \frac\gamma{\sqrt{\sigma^2+\epsilon}}
  \label{formula:10}
\end{equation}
\begin{equation}
  b = \beta +(b_{FC}-\mu)\cdot \frac\gamma{\sqrt{\sigma^2+\epsilon}}
  \label{formula:11}
\end{equation}

CBNN tailors an adaptive fusing layers protocol for various activation layer scenarios, aiming to optimize the BN layer's performance.

\subsection{Maxpooling}

\begin{figure}[htbp]
\centering
\includegraphics[width=0.75\linewidth]{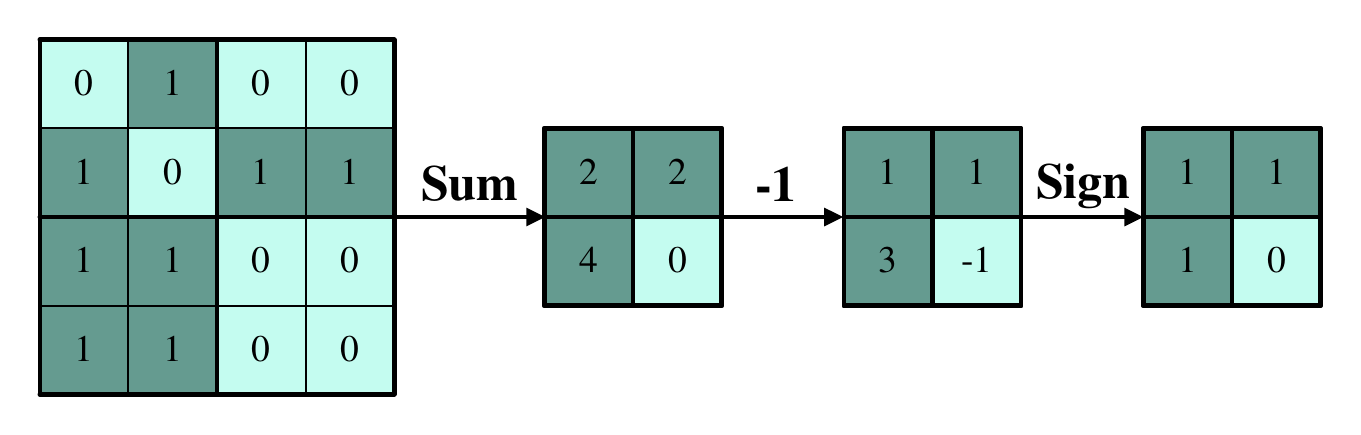}
\caption{Example of mixed protocol between the activation layer and maxpooling layer.}
\label{fig}
\end{figure}

{\color{black}Within the context of customized binary neural networks, CBNN selects the Sign function as the activation function of the preceding layer, aiming to optimize the Maxpooling operation.} 

When shares between the activation and maxpooling layers are revealed using the $Sign$ activation, the real value $X$ (either $0$ or $1$) becomes apparent, determining the maxpooling layer's output. To avoid secure compares, the maxpooling layer can fuse local addition with the Sign function. For $2 \times 2$ maxpooling with stride $2$, parties sum the shares in the window: $\left[X\right]_3^A=\left[x_0\right]_3^A+\left[x_1\right]_3^A+\left[x_2\right]_3^A+\left[x_3\right]_3^A-1$ (where 1 is a constant subtracted by one party). If the sum $X_{sum}$ is non-negative, the output is $1$ (indicating a value of $1$ in the window). Otherwise, the output is $0$. This replaces the need for secure compares, making the maxpooling layer's final step efficient with the Sign function.





\section{Experiments}
In this section, we demonstrate the practicality and efficiency of CBNN by customizing several representative BNN models and implementing secure inference. To carry out secure inference, we execute CBNN using Python, building upon the SecureBiNN framework\cite{r15}. Our experiments for secure inference are conducted on three servers provided by AutoDL, each equipped with 14 virtual CPUs and 24GB of RAM.



CBNN evaluates the benchmarks in both {\color{black}LAN and WAN} settings, providing security against semi-honest adversaries.
For the LAN setting, the average network latency is 0.2 ms, and the bandwidth is 625 MBps. In the WAN setting, the average network latency increases to 80 ms, while the bandwidth decreases to 40 MBps. 


\subsection{Evaluation on MNIST}


\begin{figure}[htbp]
\centering
\includegraphics[width=0.75\linewidth]{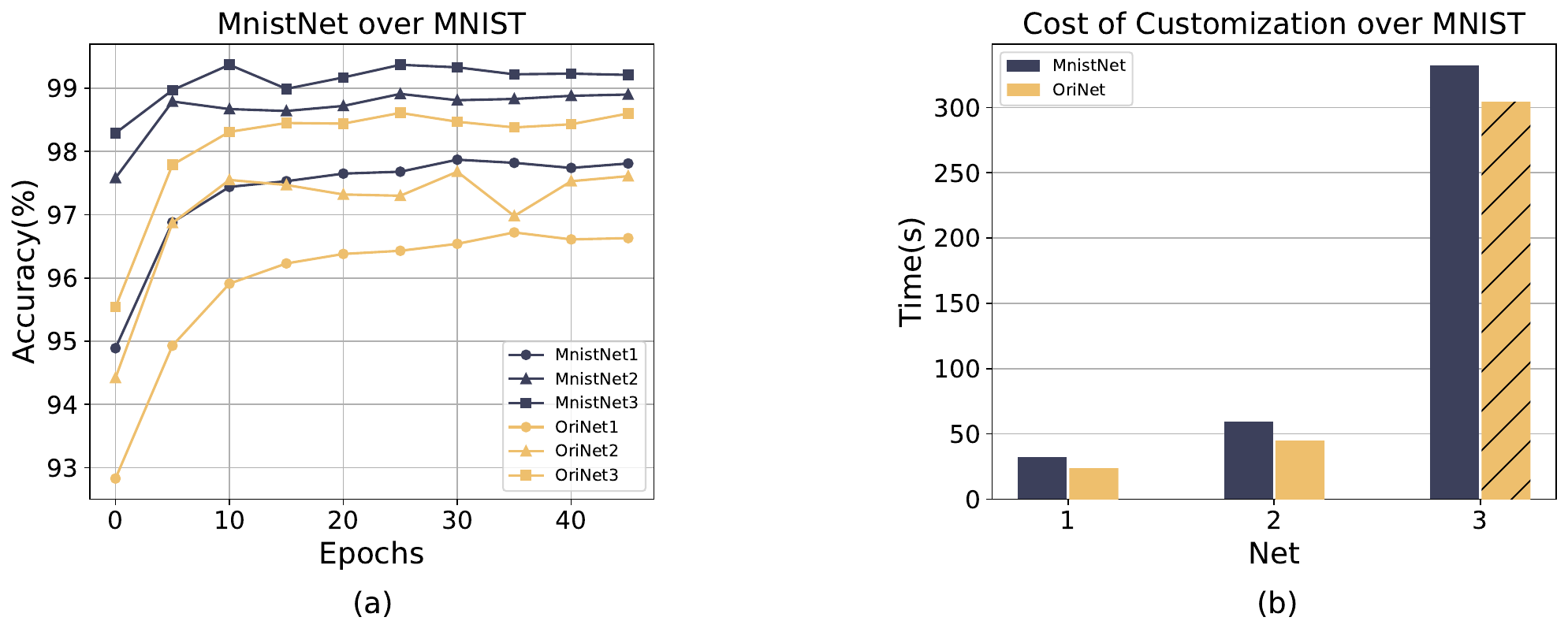}
\caption{{\color{black}Comparison of validation accuracy and training cost between typical BNNs and customized BNNs on the MNIST dataset. OriNets refer to the typically trained networks utilizing the architectures of MnistNets.}}
\label{fig:6}
\end{figure}

\textbf{Analysis of Customized BNN:} {\color{black}
To demonstrate the customization's effectiveness, we train typical BNNs without the KD algorithm, referred to as OriNets, which share identical architectures with MnistNet (described in Table \ref{tab:3}). Figure \ref{fig:6}(a) demonstrates that the KD algorithm, when employed with a highly accurate teacher model, not only facilitates faster network training but also produces superior network performance.

Furthermore, from Figure~\ref{fig:6}(b), it is evident that CBNN training exhibits reasonable complexity. While MnistNet incorporates additional steps compared to traditional training due to the KD algorithm, CBNN training accelerates convergence and tends to require fewer iterations.

\begin{table}[htbp]\small
	\caption{Evaluation Results of CBNN on MNIST and Comparisons with Prior Work under LAN/WAN Setting. 
 } 
	\label{tab:2}
	\centering
    \begin{tabular*}{\linewidth}{@{}c|c|c|c|c|c@{}}
		\toprule
        \makebox[0.15\textwidth][c]{Arch.} & \makebox[0.18\textwidth][c]{Framework}& \makebox[0.165\textwidth][c]{Time(s,LAN)}& \makebox[0.165\textwidth][c]{Time(s,WAN)} & \makebox[0.15\textwidth][c]{ Comm.(MB)}& \makebox[0.14\textwidth][c]{Acc.(\%)} \cr
		\midrule
		\multirow{6}{*}{MnistNet1}
        & ABNN2 & 1.008 & 2.44 & 4.33 & 97.6 \\
        & XoNN & 0.13 & - & 4.29 & 97.6 \\
		& SecureNN & 0.043 & 2.43 & 2.1 & 93.4 \\
		& Falcon & 0.011 & 0.99 & 0.012 & 97.4 \\
        & SecureBiNN & 0.010 & 0.248 & 0.005 & 97.3 \\
        & CBNN(ours) & 0.010 & 0.21 & 0.010 & 98.11 \\
		\cmidrule{1-6}
		\multirow{5}{*}{MnistNet2}
        & XoNN & 0.16 & - & 38.3 & 98.6 \\
		& SecureNN & 0.076 & 3.06 & 4.05 & 98.8 \\
		& Falcon & 0.009 & 0.76 & 0.049 & 97.8 \\
        & SecureBiNN & 0.007 & 0.44 & 0.032 & 97.2 \\
        & CBNN(ours) & 0.010 & 0.32 & 0.033 & 98.3 \\
		\cmidrule{1-6}
		\multirow{5}{*}{MnistNet3}
		& XoNN & 0.15 & - & 32.1 & 99.0 \\
        & SecureNN & 0.13 & 3.93 & 8.86 & 99.0 \\
		& Falcon & 0.042 & 3.0 & 0.51 & 98.6 \\
        & SecureBiNN & 0.020 & 1.15 & 0.357 & 98.4 \\
        & CBNN(ours) & 0.015 & 0.97 & 0.370 & 99.0 \\

		\bottomrule
	\end{tabular*}
\end{table}

\textbf{Comparison with prior work:\ }
This study compares the performance of CBNN with various state-of-the-art secure neural network frameworks, including SecureBiNN\cite{r15}, XONN\cite{r12}, Falcon\cite{r10}, and others\cite{r7,r51}. The comparative results are presented in Table \ref{tab:2}.

{\color{black}
As evident from Table~\ref{tab:2}, CBNN demonstrates superior performance in the LAN environment. Owing to the simplicity of MnistNets, they are not customized with MPC-friendly convolutions, resulting in additional inference time and communication overhead. Notably, the KD algorithm has a limited impact on enhancing inference accuracy given the straightforwardness of the network architectures.

However, in the WAN setting, the optimization of secure inference protocols by CBNN is further emphasized by network latency. We observe that CBNN achieves reduced communication rounds through the efficient MSB extraction protocol described in Algorithm \ref{alg:3}.} Additionally, the fusion protocol within the batch normalization layer and the efficient algorithm in the maxpooling layer also contribute to enhancing the efficiency of secure inference.

\subsection{Evaluation on CIFAR-10}


\textbf{Analysis of Customized BNN:\ }
Analogous to our analysis of the MNIST dataset, we aim to evaluate both customized BNNs and conventional BNNs to demonstrate the efficacy of customization. Network architectures (CifarNets) are described in Table \ref{tab:3}.

\begin{figure}[htbp]
\centering
\includegraphics[width=0.75\linewidth]{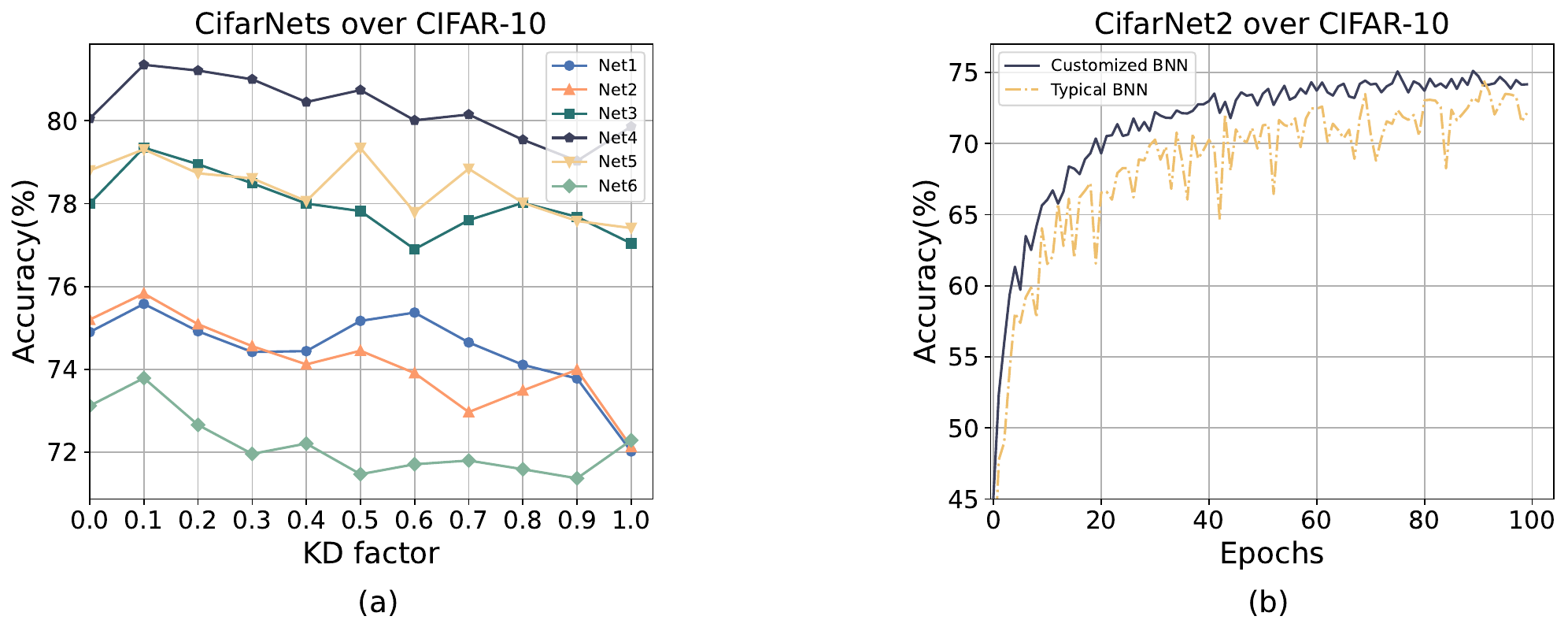}
\caption{Effect of KD weighting factor $\lambda$ of knowledge distillation in secure inference accuracy and comparison in validation accuracy for training typical BNNs and
customized BNNs over CIFAR-10.}
\label{fig:7}
\end{figure}

Figure\ref{fig:7}(a) illustrates the inference accuracy achieved with varying KD weighting factors ($\lambda$) in the KD algorithm. This weighting factor ($\lambda$) determines the extent to which the student model contributes to the loss function. As the weighting factor increases, the network's accuracy decreases until it reaches $\lambda=1$, indicating training without the KD algorithm. 

Figure\ref{fig:7}(b) demonstrates that the customization method enhances the performance of model training and expedites the convergence speed of BNN.

\begin{table}[htbp]
	\caption{The evaluation Results of CifarNet2 under LAN/WAN setting on CIFAR-10 and Comparisons with
Typical Network.} 
	\label{tab:5}
	\centering
    \begin{tabular*}{\linewidth}{@{}cccccc@{}}
		\toprule
  \makebox[0.16\textwidth][c]{Arch.} &
  \makebox[0.16\textwidth][c]{Time(s,LAN)} &
  \makebox[0.16\textwidth][c]{Time(s,WAN)} &
  \makebox[0.16\textwidth][c]{Comm.(MB)} &
  \makebox[0.16\textwidth][c]{Acc.(\%)} &
  \makebox[0.16\textwidth][c]{Para.} \cr
		\midrule
		\multirow{2}{2cm}{\centering Typical BNN\\CifarNet2}
		& 0.532 & 3.12 & 12.58 & 83.52 & 383,858 \\
        & 0.311 & 0.871 & 8.29 & 81.53 & 67,949 \\
  
  
  	\cmidrule{1-6}
		\multirow{1}{*}{Change}
		& -41.5\% & -72.1\% & -35.8\% & -1.99\% & -82.3\% \\

		\bottomrule
	\end{tabular*}
\end{table}

Given the limited number of channels (3) in CIFAR-10, CBNN employs MPC-friendly convolutions. Table~\ref{tab:5} demonstrates the performance of MPC-friendly convolutions in secure inference. We train the typical BNN using the KD algorithm, employing the same teacher model (VGG16) as CifarNet2. The results highlight the efficiency benefits of MPC-friendly convolutions, which reduce parameters to improve inference speed while sacrificing minimal accuracy.

\begin{table}[htbp]
	\caption{Evaluation Results of CBNN under LAN/WAN Setting on CIFAR-10 and Comparisons with Prior Work. } 
	\label{tab:6.1}
	\centering
    \begin{tabular*}{\linewidth}{@{}ccccc@{}}
		\toprule
		\makebox[0.2\textwidth][c]{Framework} & \makebox[0.2\textwidth][c]{Time(s,LAN)} & 
        \makebox[0.2\textwidth][c]{Time(s,WAN)}&
        \makebox[0.2\textwidth][c]{Comm.(MB)} &\makebox[0.2\textwidth][c]{Acc.(\%)} \cr
		\midrule
		MiniONN &  544 & - &  9272 & 81.61 \\
		Chameleon &   52.67 & - &  2650 & 81.61 \\
		EzPC &  265.6 & - &  40683 & 81.61 \\
        Gazelle &   15.48 & - &  1236 & 81.61 \\
        XONN &  5.79 & - &  2599 & 81.85 \\
        Falcon &  0.79 & 1.27  & 13.51 & 81.61 \\
        SecureBiNN &  0.527 & 3.447 &  16.609 &  81.50\\
        CBNN(ours) &  0.311 & 0.871 & 8.291 & 81.53 \\
		\bottomrule
      \end{tabular*}
\end{table}

\textbf{Comparison with prior work.\ }
This work compares the performance of CBNN with MiniONN\cite{r29}, Chameleon\cite{r27}, EzPC\cite{r28}, Gazelle\cite{r30}, XONN\cite{r12}, and SecureBiNN\cite{r15}, as summarized in Table~\ref{tab:6.1}. The relatively intricate model architecture of CBNN leverages its customization capabilities to its full potential. When compared to prior works under the LAN setting, our experiments on CIFAR-10 demonstrate superior performance compared to other frameworks. 

{\color{black}
Under the WAN setting, the performance of CBNN is further optimized when compared to SecureBiNN and Falcon. The efficient secure protocols employed in non-linear operations play a crucial role in minimizing the number of communication rounds. This reduction is particularly significant in a high-latency WAN environment, where the decreased communication cost directly leads to a shorter secure inference time.
}

\section{Conclusion}
In this work, we introduced the CBNN framework, a secure computation platform tailored for customized binary neural network inference. The CBNN framework introduces innovative techniques to adapt neural networks into models compatible with secure multi-party computation (MPC). By leveraging binarization and separable convolutions for model compression, MPC-friendly models achieve a significant reduction in parameters and bits, leading to reduced communication and computation costs. To offset the potential loss in inference accuracy due to simplified model architectures, we incorporated the knowledge distillation technique. CBNN effectively enhances inference accuracy while maintaining privacy. Through the development of novel secure protocols for non-linear operations, CBNN demonstrates superior performance compared to both customized and typical binary neural networks. Experimental results clearly demonstrate the applicability and efficiency of the CBNN framework.

\begin{appendix}

\section{Network Architectures}

\textbf{Network Configurations:}  In this study, we customize and
conduct secure evaluations of MnistNet1, MnistNet2, and MnistNet3 for the MNIST dataset. We train these networks using the standard BNN training algorithm\cite{r33} and further customize the training algorithm with the KD algorithm for self-comparison. 

{\color{black}MPC-friendly convolutions are not utilized in MnistNets due to the limited number of channels.} To this end, we select MnistNet4, which has an identical architecture to MnistNet3 but with more parameters, as the teacher model with a weighting factor of $\lambda=0.1$ and a temperature of $T=10$. MnistNet4 does not require MPC-friendly convolutions and uses ReLU as the activation function rather than $Sign$. Table \ref{tab:1} provides a summary of the MnistNet architectures.

For the CIFAR-10 dataset, prior works have employed six distinct network architectures. CifarNet1 represents the binary variant of the architecture introduced by MiniONN\cite{r29}. To assess the scalability of our framework to larger networks, we have binarized and customized the Fitnet\cite{r36} architectures, designated as CifarNet2 through CifarNet5. Furthermore, we evaluate the popular VGG16\cite{r37} architecture as CifarNet6. For the customized BNNs, we selected VGG16 with full-precision (32 bits) and ResNet18\cite{r38} as pre-trained teacher models. Table \ref{tab:3} provides a comprehensive overview of the architectures used for the CIFAR-10 dataset.

\begin{table}[htbp]
	\caption{Summary of the Customized Network and Teacher Model Architectures on MNIST and CIFAR-10 Dataset.} 
	
    \label{tab:1}
    \label{tab:3}
    \label{tab:7}
    
	\centering
    \begin{tabular*}{\linewidth}{@{}cccc@{}}
		\toprule
		\makebox[0.3\textwidth][c]{Arch.} & \makebox[0.3\textwidth][c]{layers} &\makebox[0.3\textwidth][c]{Description} \cr
		\midrule
		MnistNet1 & 3 & 3FC \\
		MnistNet2 &  3 &  1 CONV, 2 FC \\
		MnistNet3 &  6 & 2 CONV, 2 MP, 2 FC \\
        MnistNet4 (teacher)  & 6 & 2 CONV, 2 MP, 2 FC \\
        CifarNet1 &  10 & 7 CONV, 2 MP, 1 FC \\
		CifarNet2 &  13 & 9 CONV, 3 MP, 1 FC \\
		CifarNet3 &  13 &  9 CONV, 3 MP, 1 FC \\
        CifarNet4 &  15 & 11 CONV, 3 MP, 1 FC \\
        CifarNet5 &  21 & 17 CONV, 3 MP, 1 FC \\
        CifarNet6 & 19 &  13 CONV, 5 MP, 3 FC \\
        CifarNet7(teacher) &  19 &  13 CONV, 5 MP, 3 FC \\
        CifarNet8(teacher) &  22 & ResNet18 \\
		\bottomrule
	\end{tabular*}
\end{table}

\section{Related Work}
In recent years, significant research attention has been focused on the secure inference of privacy-preserving machine learning (PPML). Previous approaches primarily relied on secure multiparty computation (MPC) and homomorphic encryption (HE) for secure inference. However, the groundbreaking implementation of CryptoNets\cite{r3} in 2016 highlighted the substantial performance overhead associated with HE in secure inference scenarios.

With the advancement of MPC, PPML schemes leveraging MPC have demonstrated superior efficiency and accuracy compared to those relying solely on HE. Techniques utilizing secret sharing (SS) have emerged as a popular choice for PPML applications due to their attractive performance characteristics. SecureML\cite{r31}, which is based on secret sharing, employs Beaver's Triplet\cite{r43} to efficiently realize multiplication operations. Additionally, frameworks like SecureBiNN\cite{r15} and Falcon\cite{r10} adopt replicated secret sharing\cite{r20} to develop PPML solutions. MPC not only helps reduce running time and communication costs but also enables the evaluation of non-linear operations based on SS. However, this approach tends to compromise inference accuracy. To address this, frameworks like DeepSecure\cite{r44}, Chameleon\cite{r27}, XONN\cite{r12}, and SecureBiNN\cite{r15} utilize mixed protocols, incorporating Garbled Circuits (GC) to enhance the PPML framework. Frameworks built upon ABY3\cite{r5} further enhance performance by facilitating the translation of shares between Boolean and arithmetic circuits, enabling the efficient implementation of non-linear functions.

Acknowledging the constraints of PPML, customized networks tailored for specific operations consistently demonstrate superior performance in secure inference. XONN\cite{r12}, a two-party framework, effectively leverages Garbled Circuits (GC) to facilitate secure inference on vast binarized neural networks. Similarly, the three-party frameworks SecureBiNN\cite{r15} and Banners\cite{r45} adopt a hybrid approach, utilizing GC for non-linear functions and Replicated Secret Sharing (RSS) for linear computations, to implement secure inference for Binary Neural Networks (BNN). Alternatively, QUOTIENT\cite{r46}, SecureQNN\cite{r47}, and Falcon\cite{r10} introduce distinct quantization methods\cite{r1} to construct efficient PPML frameworks. This study continues this trend of research, aiming to further enhance secure inference performance by customizing in BNNs and primarily relying on the RSS technique.

\end{appendix}

\bibliography{ref}

\end{document}